\documentclass{acmart}

\usepackage{booktabs}
\usepackage{listings}

\AtBeginDocument{%
  \providecommand\BibTeX{{%
    \normalfont B\kern-0.5em{\scshape i\kern-0.25em b}\kern-0.8em\TeX}}}

\copyrightyear{2021}
\acmYear{2021}
\setcopyright{iw3c2w3}
\acmConference[WWW '21 Companion]{Companion Proceedings of the Web Conference 2021}{April 19--23, 2021}{Ljubljana, Slovenia}
\acmBooktitle{Companion Proceedings of the Web Conference 2021 (WWW '21 Companion), April 19--23, 2021, Ljubljana, Slovenia}
\acmPrice{}
\acmDOI{10.1145/3442442.3458603}
\acmISBN{978-1-4503-8313-4/21/04}

\begin{document}

\title{Plumber: A Modular Framework to Create Information Extraction Pipelines}

\author{Mohamad Yaser Jaradeh}
\affiliation{\institution{L3S Research Center, Leibniz University Hannover} 
\city{Hanover}
\country{Germany}}
\email{jaradeh@l3s.de}
\orcid{0000-0001-8777-2780}

\author{Kuldeep Singh}
\affiliation{\institution{Cerence GmbH \& Zerotha Research}
\city{Aachen}
\country{Germany}}
\email{kuldeep.singh1@cerence.com}
\orcid{0000-0002-5054-9881}

\author{Markus Stocker}
\affiliation{\institution{TIB Leibniz Information Centre for Science and Technology}
\city{Hanover}
\country{Germany}}
\email{markus.stocker@tib.eu}
\orcid{0000-0001-5492-3212}

\author{S\"oren Auer}
\affiliation{\institution{TIB Leibniz Information Centre for Science and Technology}
\city{Hanover}
\country{Germany}}
\email{auer@tib.eu}
\orcid{0000-0002-0698-2864}

\renewcommand{\shortauthors}{Jaradeh et al.}

\newcommand{\NAME}{\textsc{Plumber}}

\begin{abstract}
Information Extraction (IE) tasks are commonly studied topics in various domains of research. Hence, the community continuously produces multiple techniques, solutions, and tools to perform such tasks. However, running those tools and integrating them within existing infrastructure requires time, expertise, and resources. One pertinent task here is triples extraction and linking, where structured triples are extracted from a text and aligned to an existing Knowledge Graph (KG). In this paper, we present \NAME~, the first framework that allows users to manually and automatically create suitable IE pipelines from a community-created pool of tools to perform triple extraction and alignment on unstructured text. Our approach provides an interactive medium to alter the pipelines and perform IE tasks. A short video to show the working of the framework for different use-cases is available online\footnote{\url{https://www.youtube.com/watch?v=XC9rJNIUv8g}}
\end{abstract}




\maketitle

\section{Introduction}
\begin{figure*}[t]
	\centering
	\includegraphics[width=.78\textwidth]{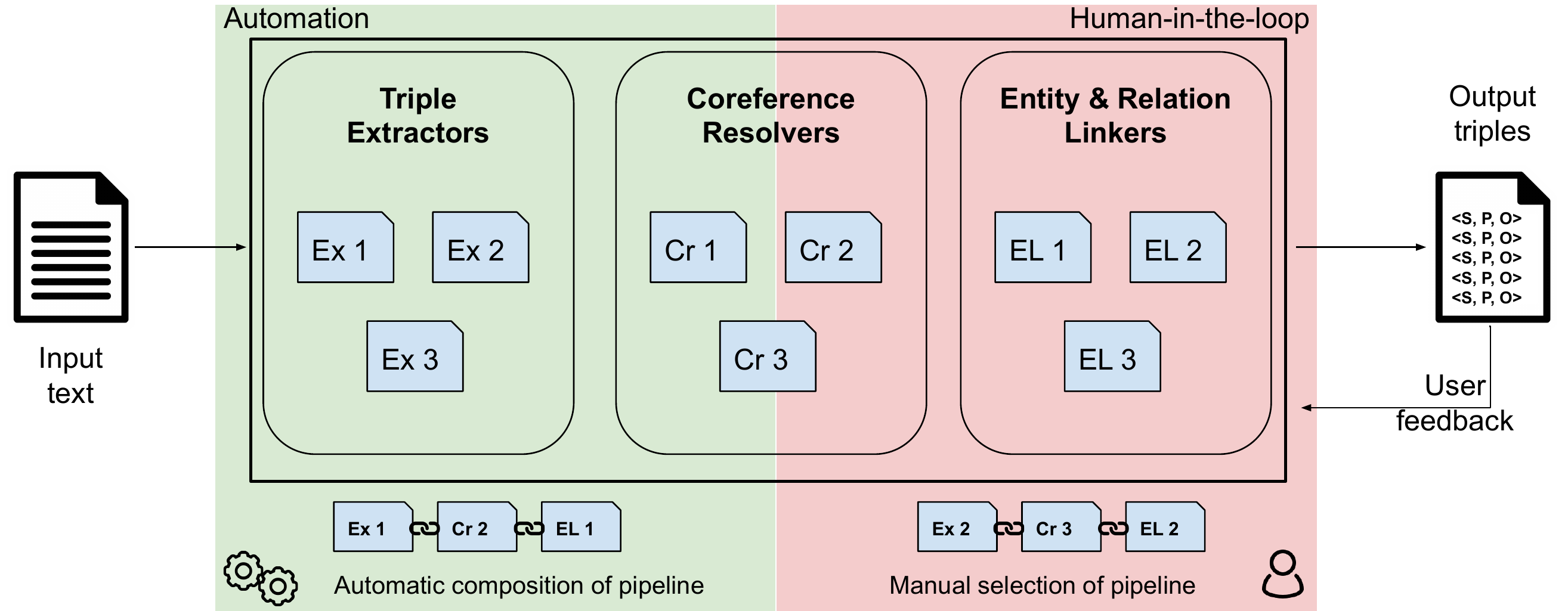}
	\caption{\NAME's schematics showing the interactions between users and the system for use-case illustrations. Users can select pipelines manually and provide feedback on resulting triples. While, the system can compose a pipeline based on the input text automatically.
	}
	\label{fig:demo}
\end{figure*}

Continuous efforts has been made in the Web community since the early 21st century to extend the Web as a global data graph using RDF~\cite{berners2001semantic}. Such efforts produced valuable linked resources that are prominent on the Web such as DBpedia~\cite{dbpedia} and YAGO~\cite{fabian2007yago} Knowledge Graphs (KGs). The Web research community has used these KGs in Information Extraction (IE) tasks such as triple extraction~\cite{kgBert}, keywords and topics extraction~\cite{textflow}, to entities \& relations extraction and linking~\cite{falcon}.
Researchers compose information extraction pipelines via chaining the aforementioned tasks together to perform a variety of applications such as question answering, KG completion, fact checking, and dialog systems~\cite{kgBert}.
We demonstrate \NAME, a Web-based tool that consolidates the community efforts by bringing in various open-sourced and online-available tools under one umbrella. \NAME~ is the core implementation of a methodology (see \autoref{fig:demo}) which is grounded in three principles
1) Reusability: the framework is open source and reusable that includes a web-based UI for choosing and integrated components.
2) Isolation: all IE components implemented under \NAME~ operate in isolation of each other.
3) Extensibility: the framework is extensible to new components and other variation of pipelines.

\NAME~ distinguishes itself from other pipelining frameworks by allowing two modes of operation. 
1) Manual: a user is able to select the components composing the resulting pipeline by hand.
2) Automatic: the framework makes use of contextual embeddings to automatically compose a suitable pipeline for a given input text.

For wider adaptation of \NAME~, it is integrated within the Open Research Knowledge Graph (ORKG) infrastructure~\cite{orkg} throughout its user interface. The interface allows users to provide text snippets or files as input and allows \NAME~ to compose a suitable pipeline based on the provided input to produce the final set of triple statements extracted from the input text.

\section{\NAME}

\NAME~ is a modular framework that integrates 40 information extraction components. The framework relies on the principle of Isolation, Extensibility, and Reusability. Like other counterparts in the Question Answering (QA) domain (i.e., Frankenstein~\cite{frankenstein}), \NAME~ also composes suitable pipelines based on user input. Frankenstein relies on logistic regression-based pipeline selection. However, \NAME~ we use RoBERTa based classifier \cite{liu2019roberta}, which is trained on micro F-score of each pipeline executed end-to-end. In the Information Extraction community, no such effort has been made other than \NAME~. Our proposed framework re-uses an ontology from Frankenstein for solving data interoperability issues of various integrated components. 
\NAME~ abstracts the details of creating IE pipelines and has two modes of operation: Manual and Automatic, depending on IE pipeline selection.
These IE pipelines are composed of the following IE tasks 
1) Text Triple Extraction: firstly, a set of text triple needs to be extracted from the input text snippets.
2) Coreference Resolution: all mentions and pronouns are resolved and replaced with their original mention.
3) Entity and Relation Linking: these tools extract the surface form of entities and predicates in an unstructured text and link it to their corresponding KG mentions. 
The implementation of \NAME~ is available and released on GitHub\footnote{\url{https://github.com/YaserJaradeh/ThePlumber}}. For a detailed empirical evaluation on effectiveness of pipeline composition, we refer readers to ~\cite{jaradeh2021better}.
\NAME's approach for creating pipelines is as follows:

\textbf{Pipeline Pool Population} --- \NAME~ has an internal repository of candidate components that implement the three IE tasks mentioned earlier. \NAME~ generates all possible pipelines from the underlying pool and tags them with specific characteristics (i.e., which KG they align to) and adds them to its pipelines pool.

\textbf{Pipeline Composer} --- The user can guide the framework's selection process (i.e., components of the pipeline are user specified). A RoBERTa model selects the most suitable pipeline from the pipeline pool that can produce the best results out of the input text.

\textbf{Pipeline Runner} --- Once a pipeline is selected, the framework then instantiates the pipeline, passes the input text specified by the user, and waits for the results (i.e., aligned triples), which are then displayed for users via the user interface.

\textbf{Triples Feedback} --- Considering that the aforementioned process is automatic, once the user can see the triples via the UI, they can report incorrect triples that feedback into the framework to improve accuracy.

\section{Demonstration}
In this demonstration, we will show \NAME~ in action, highlight the two different use cases of selecting an IE pipeline manually, and let the framework choose a suitable one based on the input text's characteristics. 
The demonstration transitions step by step from providing the input text and then showing the two different use cases of manual and automatic pipeline selection to comparing the resulting final set of triples. We believe our work to connect disjoint IE efforts on the Web will motivate the researchers to provide more efficient components complementing other domains' efforts.

\begin{acks}
This work was co-funded by the European Research Council for the project Science GRAPH (Grant agreement ID: 819536) and the TIB Leibniz Information Centre for Science and Technology. 
\end{acks}

\bibliographystyle{ACM-Reference-Format}
\bibliography{references}


\begin{thebibliography}{10}


\ifx \showCODEN    \undefined \def \showCODEN     #1{\unskip}     \fi
\ifx \showDOI      \undefined \def \showDOI       #1{#1}\fi
\ifx \showISBNx    \undefined \def \showISBNx     #1{\unskip}     \fi
\ifx \showISBNxiii \undefined \def \showISBNxiii  #1{\unskip}     \fi
\ifx \showISSN     \undefined \def \showISSN      #1{\unskip}     \fi
\ifx \showLCCN     \undefined \def \showLCCN      #1{\unskip}     \fi
\ifx \shownote     \undefined \def \shownote      #1{#1}          \fi
\ifx \showarticletitle \undefined \def \showarticletitle #1{#1}   \fi
\ifx \showURL      \undefined \def \showURL       {\relax}        \fi
\providecommand\bibfield[2]{#2}
\providecommand\bibinfo[2]{#2}
\providecommand\natexlab[1]{#1}
\providecommand\showeprint[2][]{arXiv:#2}

\bibitem[\protect\citeauthoryear{Auer, Bizer, Kobilarov, Lehmann, Cyganiak, and
  Ives}{Auer et~al\mbox{.}}{2007}]%
        {dbpedia}
\bibfield{author}{\bibinfo{person}{S{\"o}ren Auer}, \bibinfo{person}{Christian
  Bizer}, \bibinfo{person}{Georgi Kobilarov}, \bibinfo{person}{Jens Lehmann},
  \bibinfo{person}{Richard Cyganiak}, {and} \bibinfo{person}{Zachary Ives}.}
  \bibinfo{year}{2007}\natexlab{}.
\newblock \showarticletitle{DBpedia: A Nucleus for a Web of Open Data}. In
  \bibinfo{booktitle}{\emph{The Semantic Web}}. \bibinfo{publisher}{Springer
  Berlin Heidelberg}, \bibinfo{pages}{722--735}.
\newblock


\bibitem[\protect\citeauthoryear{Berners-Lee, Hendler, and Lassila}{Berners-Lee
  et~al\mbox{.}}{2001}]%
        {berners2001semantic}
\bibfield{author}{\bibinfo{person}{Tim Berners-Lee}, \bibinfo{person}{James
  Hendler}, {and} \bibinfo{person}{Ora Lassila}.}
  \bibinfo{year}{2001}\natexlab{}.
\newblock \showarticletitle{The semantic web}.
\newblock \bibinfo{journal}{\emph{Scientific american}} \bibinfo{volume}{284},
  \bibinfo{number}{5} (\bibinfo{year}{2001}), \bibinfo{pages}{34--43}.
\newblock


\bibitem[\protect\citeauthoryear{{Cui}, {Liu}, {Tan}, {Shi}, {Song}, {Gao},
  {Qu}, and {Tong}}{{Cui} et~al\mbox{.}}{2011}]%
        {textflow}
\bibfield{author}{\bibinfo{person}{W. {Cui}}, \bibinfo{person}{S. {Liu}},
  \bibinfo{person}{L. {Tan}}, \bibinfo{person}{C. {Shi}}, \bibinfo{person}{Y.
  {Song}}, \bibinfo{person}{Z. {Gao}}, \bibinfo{person}{H. {Qu}}, {and}
  \bibinfo{person}{X. {Tong}}.} \bibinfo{year}{2011}\natexlab{}.
\newblock \showarticletitle{TextFlow: Towards Better Understanding of Evolving
  Topics in Text}.
\newblock \bibinfo{journal}{\emph{IEEE Transactions on Visualization and
  Computer Graphics}} \bibinfo{volume}{17}, \bibinfo{number}{12}
  (\bibinfo{year}{2011}), \bibinfo{pages}{2412--2421}.
\newblock


\bibitem[\protect\citeauthoryear{Fabian, Gjergji, Gerhard,
  et~al\mbox{.}}{Fabian et~al\mbox{.}}{2007}]%
        {fabian2007yago}
\bibfield{author}{\bibinfo{person}{MS Fabian}, \bibinfo{person}{Kasneci
  Gjergji}, \bibinfo{person}{WEIKUM Gerhard}, {et~al\mbox{.}}}
  \bibinfo{year}{2007}\natexlab{}.
\newblock \showarticletitle{Yago: A core of semantic knowledge unifying wordnet
  and wikipedia}. In \bibinfo{booktitle}{\emph{16th International World Wide
  Web Conference, WWW}}. \bibinfo{pages}{697--706}.
\newblock


\bibitem[\protect\citeauthoryear{Jaradeh, Oelen, Farfar, Prinz, D'Souza,
  Kismih{\'{o}}k, Stocker, and Auer}{Jaradeh et~al\mbox{.}}{2019}]%
        {orkg}
\bibfield{author}{\bibinfo{person}{Mohamad~Yaser Jaradeh},
  \bibinfo{person}{Allard Oelen}, \bibinfo{person}{Kheir~Eddine Farfar},
  \bibinfo{person}{Manuel Prinz}, \bibinfo{person}{Jennifer D'Souza},
  \bibinfo{person}{Gábor Kismih{\'{o}}k}, \bibinfo{person}{Markus Stocker},
  {and} \bibinfo{person}{Sören Auer}.} \bibinfo{year}{2019}\natexlab{}.
\newblock \showarticletitle{{Open Research Knowledge Graph: Next Generation
  Infrastructure for Semantic Scholarly Knowledge}}.
\newblock \bibinfo{journal}{\emph{Marina Del K-CAP}}  \bibinfo{volume}{19}
  (\bibinfo{year}{2019}).
\newblock


\bibitem[\protect\citeauthoryear{Jaradeh, Singh, Stocker, Both, and
  Auer}{Jaradeh et~al\mbox{.}}{2021}]%
        {jaradeh2021better}
\bibfield{author}{\bibinfo{person}{Mohamad~Yaser Jaradeh},
  \bibinfo{person}{Kuldeep Singh}, \bibinfo{person}{Markus Stocker},
  \bibinfo{person}{Andreas Both}, {and} \bibinfo{person}{S{\"o}ren Auer}.}
  \bibinfo{year}{2021}\natexlab{}.
\newblock \showarticletitle{Better Call the Plumber: Orchestrating Dynamic
  Information Extraction Pipelines}.
\newblock \bibinfo{journal}{\emph{International Conference on Web Engineering
  (ICWE)}} (\bibinfo{year}{2021}).
\newblock


\bibitem[\protect\citeauthoryear{Liu, Ott, Goyal, Du, Joshi, Chen, Levy, Lewis,
  Zettlemoyer, and Stoyanov}{Liu et~al\mbox{.}}{2019}]%
        {liu2019roberta}
\bibfield{author}{\bibinfo{person}{Yinhan Liu}, \bibinfo{person}{Myle Ott},
  \bibinfo{person}{Naman Goyal}, \bibinfo{person}{Jingfei Du},
  \bibinfo{person}{Mandar Joshi}, \bibinfo{person}{Danqi Chen},
  \bibinfo{person}{Omer Levy}, \bibinfo{person}{Mike Lewis},
  \bibinfo{person}{Luke Zettlemoyer}, {and} \bibinfo{person}{Veselin
  Stoyanov}.} \bibinfo{year}{2019}\natexlab{}.
\newblock \showarticletitle{Roberta: A robustly optimized bert pretraining
  approach}.
\newblock \bibinfo{journal}{\emph{preprint arXiv:1907.11692}}
  (\bibinfo{year}{2019}).
\newblock


\bibitem[\protect\citeauthoryear{Sakor, Onando~Mulang{'}, Singh, Shekarpour,
  Esther~Vidal, Lehmann, and Auer}{Sakor et~al\mbox{.}}{2019}]%
        {falcon}
\bibfield{author}{\bibinfo{person}{Ahmad Sakor}, \bibinfo{person}{Isaiah
  Onando~Mulang{'}}, \bibinfo{person}{Kuldeep Singh}, \bibinfo{person}{Saeedeh
  Shekarpour}, \bibinfo{person}{Maria Esther~Vidal}, \bibinfo{person}{Jens
  Lehmann}, {and} \bibinfo{person}{S{\"o}ren Auer}.}
  \bibinfo{year}{2019}\natexlab{}.
\newblock \showarticletitle{Old is Gold: Linguistic Driven Approach for Entity
  and Relation Linking of Short Text}. \bibinfo{publisher}{Association for
  Computational Linguistics}, \bibinfo{pages}{2336--2346}.
\newblock


\bibitem[\protect\citeauthoryear{Singh, Radhakrishna, Both, Shekarpour, Lytra,
  Usbeck, Vyas, Khikmatullaev, Punjani, Lange, Vidal, Lehmann, and Auer}{Singh
  et~al\mbox{.}}{2018}]%
        {frankenstein}
\bibfield{author}{\bibinfo{person}{Kuldeep Singh},
  \bibinfo{person}{Arun~Sethupat Radhakrishna}, \bibinfo{person}{Andreas Both},
  \bibinfo{person}{Saeedeh Shekarpour}, \bibinfo{person}{Ioanna Lytra},
  \bibinfo{person}{Ricardo Usbeck}, \bibinfo{person}{Akhilesh Vyas},
  \bibinfo{person}{Akmal Khikmatullaev}, \bibinfo{person}{Dharmen Punjani},
  \bibinfo{person}{Christoph Lange}, \bibinfo{person}{Maria~Esther Vidal},
  \bibinfo{person}{Jens Lehmann}, {and} \bibinfo{person}{S\"{o}ren Auer}.}
  \bibinfo{year}{2018}\natexlab{}.
\newblock \showarticletitle{Why Reinvent the Wheel: Let's Build Question
  Answering Systems Together}. In \bibinfo{booktitle}{\emph{Proceedings of the
  2018 World Wide Web Conference}} \emph{(\bibinfo{series}{WWW '18})}.
  \bibinfo{pages}{1247–1256}.
\newblock


\bibitem[\protect\citeauthoryear{Yao, Mao, and Luo}{Yao et~al\mbox{.}}{2019}]%
        {kgBert}
\bibfield{author}{\bibinfo{person}{Liang Yao}, \bibinfo{person}{Chengsheng
  Mao}, {and} \bibinfo{person}{Yuan Luo}.} \bibinfo{year}{2019}\natexlab{}.
\newblock \bibinfo{title}{KG-BERT: BERT for Knowledge Graph Completion}.
\newblock
\newblock
\showeprint[arxiv]{cs.CL/1909.03193}


\end{thebibliography}

\end{document}